\documentclass[a4paper]{article}

\usepackage{INTERSPEECH2021}
\usepackage{amsmath,graphicx}
\usepackage{boxedminipage}
\usepackage{wrapfig,lipsum,booktabs}
\usepackage{times}
\usepackage{longtable}
\usepackage{amsmath}
\usepackage{subcaption}
\usepackage{tabularx}
\usepackage{colortbl}
\usepackage{latexsym}
\usepackage{makecell}
\usepackage{graphicx}
\usepackage{caption}
\usepackage{multirow}
\usepackage{makecell}

\usepackage{microtype}

\newcommand{\B}[1] {\boldsymbol{#1}}

\def\bm{{\B{m}}}

\def\bx{{\B{x}}}

\newcommand{\Acal}{\mathcal{A}}
\newcommand{\Dcal}{\mathcal{D}}
\newcommand{\Imat}{{\bf I}}

\makeatletter
\newcommand{\thickhline}{%
    \noalign {\ifnum 0=`}\fi \hrule height 1pt
    \futurelet \reserved@a \@xhline
}

\newcommand{\thickerhline}{%
    \noalign {\ifnum 0=`}\fi \hrule height 2pt
    \futurelet \reserved@a \@xhline
}

\makeatother

\title{Data Augmentation for Spoken Language Understanding via Pretrained Language Models}
\name{Baolin Peng$^*$\thanks{\hspace{0.2cm}$^*$Equal contribution}, Chenguang Zhu$^*$, Michael Zeng, Jianfeng Gao}
\address{Microsoft Research, Redmond}
\email{\{bapeng,chezhu,nzeng,jfgao\}@microsoft.com}
\begin{document}
\maketitle
\begin{abstract}
The training of spoken language understanding (SLU) models often faces the problem of data scarcity. In this paper, we put forward a data augmentation method using pretrained language models to boost the variability and accuracy of generated utterances. Furthermore, we investigate and propose solutions to two previously overlooked semi-supervised learning scenarios of data scarcity in SLU: i) \textit{Rich-in-Ontology}: ontology information with numerous valid dialogue acts is given; ii) \textit{Rich-in-Utterance}: a large number of unlabelled utterances are available. Empirical results show that our method can produce synthetic training data that boosts the performance of language understanding models in various scenarios.
\end{abstract}


\section{Introduction}
\label{sec:intro}
Spoken Language Understanding (SLU) is widely applied in human-machine dialogue systems to convert natural utterances into predefined semantic frames, i.e. dialogue acts, for further processing. For example, an SLU component in a virtual assistant or robot outputs its prediction of intents and slot labels detected within a user's utterance \cite{virtualassistant}.
Nevertheless, as a supervised learning task, SLU suffers from the problem of data scarcity. The problem becomes more prevalent in face of new LU domains with novel definitions of intents and slot labels. Even with an existing domain, the data correlated with a certain intent or slot is often not sufficient. These problems significantly limit the applicability of SLU systems.

Recently, various successful use cases of synthetic datasets have stimulated the growth of the area of Data Augmentation (DA) \cite{lu2006enhancing,hannun2014deep}. The typical approach is to learn a model to mimic the language style in the training data, leveraging the relationship between semantic units and their natural representations. Then, a non-generative model can modify utterances and replace slot labels from existing data \cite{ungen-and-gen}, while a generative model can produce synthetic utterances in the same distribution space of the training data \cite{hou-da}. However, these approaches usually train the DA model on domain-specific data, which is of a small scale by itself. It is thus questionable whether the augmented data contains rich language expressibility beyond the scope of the given data.

On the other hand, the rapid development of large-scale pretrained language models has significantly improved the capacity of language understanding and generation models \cite{bert,roberta}. 
With a modest amount of domain-specific data, a pretrained model can quickly adapt to a new domain. For instance, SC-GPT \cite{scgpt} finetunes the GPT-2 language model \cite{gpt} with dialogue data. It can efficiently adapt to new dialogue domains with only a couple of labelled data samples.

In this paper, we propose to frame data augmentation as a semantically controlled generation problem. Given dialogue act, we leverage the pretrained SC-GPT model to generate corresponding utterances as synthetic training data. In the process, the general language syntax and semantics learned during the pretraining phase are fused into the generation of domain-specific utterances to increase variability and accuracy of SLU. 

Furthermore, previous literature on SLU data augmentation focus on the case where only a scant number of pairs of utterance and corresponding semantic labels are given, which we denote as \textit{Paired-Data-Only}. However, there are two other overlooked semi-supervised learning scenarios that commonly arise in application. 

\begin{itemize}
    \item \textit{Rich-in-Ontology}: The full ontology for the dialogue domain is also given, including the definitions of intents, slot lists and valid combinations of slots and values. In other words, the model is given a variety of valid combinations of semantic labels. What lacks is the corresponding natural language utterances. 
    \item \textit{Rich-in-Utterance}: Apart from the labelled data, there are abundant unlabelled utterances without annotated intents, slots and values.
\end{itemize}

In this paper, we also delve into these two scenarios and propose corresponding data augmentation solutions.

For \textit{Rich-in-Ontology}, we first finetune the pretrained model SC-GPT on the paired training data, and then apply it to the valid combination of intents and slots in the ontology information to generate additional training data. 

For \textit{Rich-in-Utterance}, following the idea of the NLG model SC-GPT, we propose SC-GPT-NLU, which is pretrained on the same corpus of SC-GPT with flipped sources and targets. In detail, we feed the utterances into the model and let it generate intent and slots in a text sequence. Therefore, SC-GPT-NLU can act as a language understanding module and produce semantic labels for the unlabelled utterances available.

In the experiments, we evaluate the slot tagging and intent classification accuracies of a Bi-LSTM seq2seq SLU model, using various data augmentation methods. Results show that on ATIS and Snips datasets, our proposed method outperforms other baseline systems. For instance, compared with baseline methods, the data augmented by our system can help the underlying SLU model achieve 0.5 points higher slot F1 and 3.02 points higher intent accuracy in ATIS-Small dataset.
Furthermore, when ontology information or unlabelled utterances are available, i.e. Rich-in-Ontology and Rich-in-Utterance, our method can produce synthetic data that significantly boosts the performance of SLU models.




\section{Related Work}
\label{sec:related}
\subsection{SLU Data Augmentation}
Many previous approaches to SLU data augmentation target to increase variability of generated utterances. \cite{perturb} proposes to add noise to perturb the decoder states to generate variants of an utterance. Variational autoencoder (VAE) and conditional variational
autoencoder (CVAE) are used to generate utterances with diversified expressions \cite{vae-da1}. 
\cite{ungen-and-gen} uses both non-generative models like word substitution and generative models like paraphrasing and back-translation to augment training data.
\cite{hou-da} proposes a multi-stage framework to generate, filter, and rank augmented utterances.
\cite{dst-da-rl} uses reinforcement learning to learn a generator that facilitates dialogue state tracking.
\cite{atomictemplates} employs atomic templates to guide the model to generate more utterances given combination of dialogue acts. 
\cite{cho2019efficient} proposes to select sentences from unlabeled utterances and apply pseudo-labels.
The two additional scenarios we propose in this paper are also related to semi-supervised learning \cite{semisupervise}. But we focus on data augmentation, which is independent of the downstream learning models.

Similar to our work, \cite{anaby2020aaai,kumar2020data} uses pretrained language models to generate synthetic training data for data augmentation. However, their approach blends multiple labels and input sentences together during training, so it is hard to control the amount of generated synthetic data per class.

\subsection{Pretraining}
Pretrained models leverage the large amount of unlabelled text corpora to improve the capability of language understanding. 
ELMo \cite{elmo} applies two unidirectional RNNs for language modeling.
GPT-2 \cite{gpt} utilizes the transformer architecture \cite{transformer} for the task. 
BERT \cite{bert} employs a masking technique and next-sentence-prediction task to train a bidirectional language model. 
UniLM \cite{unilm} uses different masking patterns to unify the model structure for NLU and NLG. 
These pretrained language models have been widely used with considerable success in various NLP applications such as question answering \cite{sdnet} and summarization \cite{bertsum}. 

Furthermore, pretrained language models have been leveraged in speech language processing to provide rich contextual embeddings \cite{chung2020semi}. Specifically, SC-GPT \cite{scgpt}, i.e. Semantically Conditioned Generative Pre-training Transformer, builds upon GPT-2 and is further pretrained on a large-scale dialogue corpus. The resulting model outperforms many baselines in few-shot language generation for task-oriented dialogue.

\section{Data Augmentation}
\label{sec:model}

\begin{figure*}[!ht]
\centering
\vspace{-3.6cm}
\includegraphics[width=1\linewidth]{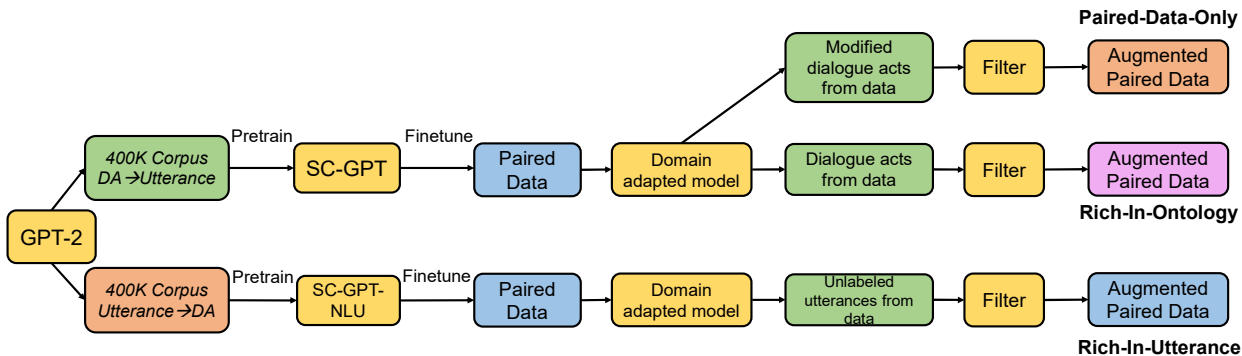}  
\vspace{-4.7cm}
\caption{Data augmentation process for the three scenarios: three Paired-Data-Only, Rich-In-Ontology and Rich-In-Utterances. All models are initialized with GPT-2, further pretrained on 400K dialogue corpus \cite{scgpt} and finetuned on the paired data $\{(\bx_1, \Acal_1), ..., (\bx_N, \Acal_N)\}$.}
\label{fig:da}
\end{figure*}

\subsection{Traditional Augmentation Scenario}
We describe the traditional augmentation scenario in SLU as \textbf{Paired-Data-Only}, as the training data consists of $N$ instance pairs. Each pair contains the input tokenized utterance $\bx=(x_1, x_2, ..., x_T)$ and the corresponding output \textit{dialogue act} $\Acal$. $\Acal$ includes the intent label $\Imat$ and $P$ slot-value pairs:
\begin{equation}
    \Acal = [ \underbrace{~~\Imat_{~_{~}}}_{\text{Intent}}, \underbrace{(s_1=v_1, \cdots, s_P= v_P)}_{\text{Slot-value pairs} } ]
\end{equation}


Thus, the training data $\Dcal=\{(\bx_1, \Acal_1), ..., (\bx_N, \Acal_N)\}$. However, due to high labeling costs, the size of labeled data $N$ is usually small. In such cases, data augmentation (DA) is needed. An augmenter $\mathbf{S}$ is a language generation model, which is trained on $\Dcal$ to be able to produce a corresponding utterance $\Tilde{x}$ given an input dialogue act $\Tilde{\Acal}$. For example, suppose $\Tilde{\Acal}=[${\it hotel-inform},$(name=Hyatt, area=center, star=5]$, $\mathbf{S}$ can generate $\Tilde{x}=${\it I have booked the 5-star Hyatt hotel in the center area for you.} 

Then, during augmentation, we first augment the dialogue acts in the training data by replacing/inserting/deleting slot values to create more combinations. The augmenter $\mathbf{S}$ then generates candidate utterances for the dialogue acts. As the generated utterances may not always contain the required slot-value labels, we filter them to make sure that each utterance has all the required input slot-values. 

However, the data augmenter itself requires a considerable amount of training data. As a result, augmenters directly trained on $\Dcal$ may have limited model capacity and expressibility. Thus, we adopt the pretrained model SC-GPT \cite{scgpt}, which is a language model to produce utterances given a dialogue act. SC-GPT is initialized with GPT-2 \cite{gpt}, further pretrained on a large corpus of 400K dialogue-act-utterance pairs and then fine-tuned on the training data $\Dcal$. It has been shown that SC-GPT can quickly adapt to new domains with only a few domain-specific data samples \cite{scgpt}.


\begin{table}[t]
\centering
    \begin{tabular}{l|l|l}
    \thickhline
    \textbf{Model} & \textbf{Input} & \textbf{Output}\\
    \hline
    SC-GPT & Dialogue act & Utterance\\
    SC-GPT-NLU & Utterance & Dialogue act\\
    \thickhline
    \end{tabular}
\caption{\label{tab:scgpt} The input and output of SC-GPT \cite{scgpt} and SC-GPT-NLU models. Both are initialized with GPT-2 \cite{gpt} but further pretrained on different data with swapped inputs and outputs.} 
\end{table} 

\subsection{More Data Augmentation Scenarios}
We note that in many real applications, there is often additional available information beyond the paired training data. Here, we specify two semi-supervised scenarios that commonly arise in applications but have been overlooked by previous approaches.

\subsubsection{Rich In Ontology}
In many dialogue domains, a detailed description of the ontology is given, which is a list of valid dialogue acts. Formally, the training data consists of both labelled pairs and many dialogue acts: $\Dcal=\{(\bx_1, \Acal_1), ..., (\bx_N, \Acal_N), \Acal_{N+1}, ..., \Acal_M\}$.

To work with this scenario, we finetune SC-GPT on the paired part of $\Dcal$, i.e. $\{(\bx_1, \Acal_1), ..., (\bx_N, \Acal_N)\}$, and then generate utterances for the other dialogue acts $\{\Acal_{N+1}, ..., \Acal_M\}$. The utterances are then filtered to make sure that each utterance has all the corresponding slot-values.


\subsubsection{Rich In Utterance}
It is common in practice that a large number of unlabelled dialogue utterances are available, usually collected from history data. Formally, the training data consists of both labelled pairs and many unlabeled utterances: $\Dcal=\{(\bx_1, \Acal_1), ..., (\bx_N, \Acal_N), \bx_{N+1}, ..., \bx_M\}$. 

\begin{table*}[ht]
\centering
\vspace{0.5cm}
    \begin{tabular}{c|l|l|l|l|l|l|l|l} \thickhline
    \textbf{Dataset} & \multicolumn{4}{c|}{ATIS} & \multicolumn{4}{c}{Snips} \\
     \hline
     \textbf{Split} & \multicolumn{2}{c|}{Small} & \multicolumn{2}{c|}{Medium} & \multicolumn{2}{c|}{Small} &
     \multicolumn{2}{c}{Medium}\\
     \hline
     \textbf{Model} & Slot F1& Intent Acc. & Slot & Intent & Slot & Intent & Slot & Intent \\
        \hline
        \rowcolor[gray]{0.95}\multicolumn{9}{c}{\textit{No Data Augmentation}}\\
        \hline
        No-DA & 68.91 & 84.99 &	87.30 & 90.15 & 61.30 & 93.43 & 79.83 & 97.29 \\
     \hline
       \rowcolor[gray]{0.95}\multicolumn{9}{c}{\textit{Paired-Data-Only}}\\
       \hline
       Seq2Seq & 73.71	& - & 88.72 & - & -& -& -& -\\
       VAE & 74.92 & 83.65 & \underline{89.27} &	\underline{90.95} & -& -& -& -\\
       Ours& \underline{75.42}	& \underline{86.67}	& 88.61	& 90.71 & 64.96 & 93.43 & 80.62 & 97.57\\
       \hline
       \rowcolor[gray]{0.95}\multicolumn{9}{c}{\textit{Rich-in-Ontology}}\\
       \hline
       Ours& \textbf{82.42$^*$} & \textbf{89.03$^*$} & \textbf{89.81$^*$} & \textbf{92.27$^*$} & \textbf{67.06$^*$} & \textbf{94.14$^*$}& \textbf{82.54$^*$}& 97.86\\
       \hline
       \rowcolor[gray]{0.95}\multicolumn{9}{c}{\textit{Rich-in-Utterance}}\\
       \hline
       Ours & 78.45 & 87.46 & 88.23 &	91.94 & 63.46 & 93.43& 80.54& \textbf{98.14$^*$}\\
       \hline
    \end{tabular}
\vspace{0.3cm}
\caption{\label{tab:main} Slot F1 and intent accuracy scores on ATIS and Snips dataset. The overall highest score is in bold, and the best result in Paired-Data-Only category is underlined. *: Statistically significant with p-value less than 0.05.} 
\end{table*} 

To utilize these utterances, we need to produce corresponding dialogue acts. We propose to finetune GPT-2 in the reverse way: feed an utterance as input and let the model generate the dialogue act as output. In other words, we leverage a language generation model to act as a language understanding module, denoted as SC-GPT-NLU (Table~\ref{tab:scgpt}).

Like SC-GPT, SC-GPT-NLU is initialized with GPT-2 and further pretrained on the 400K dialogue-act-utterance data and finetuned on the paired part of $\Dcal$. But SC-GPT-NLU treats the utterance as input and dialogue acts as output. So both SC-GPT and SC-GPT-NLU are language generation models with a softmax-based output layer that produces utterance/dialogue acts token by token.

During augmentation, SC-GPT-NLU generates dialogue acts for the unlabeled utterances $\bx_{N+1}, ..., \bx_M$. Here, the generated names of intents, slots and values are mapped to the pre-defined ontology by string matching. The augmented data is filtered to make sure that each input slot-value appears in the utterance.

Figure~\ref{fig:da} illustrates our SLU data augmentation process for all three scenarios.

\section{Experiments}
\label{sec:exp}

\subsection{Datasets and Metrics}
We employ the widely used SLU benchmark dataset ATIS \cite{atis} and Snips \cite{snips}. ATIS contains around 5.8K utterances from flight reservation dialogues. It includes 120 slot labels and 21 intent types. Snips contains 14K utterances from the Snips personal voice assistant. It includes 123 slot labels and 7 intent types.

To simulate the few-shot data situations, we follow \cite{datasplit} to use two small portions of the ATIS training set as training data: Small ($\sim$1/40 of the original training set) and Medium ($\sim$1/10 of the original training set). A development set of 500 instances is used. Following the same split ratio, we sampled 327 and 1308 instances in Snips for Small and Medium respectively. 

We use F1 score to measure slot tagging quality and use accuracy score to evaluate intent classification, in accordance with \cite{datasplit}.

\begin{table*}[tbp]
\centering
    \begin{tabular}{c|l} \thickhline
       \rowcolor[gray]{0.95}\multicolumn{2}{c}{\textit{SC-GPT}}\\
       \hline
       \textbf{DA} & RateBook (best\_rating = 6; object\_select = current; object\_type = textbook; rating\_value = 3) \\
       \hline
       \textbf{Utterance 1} & Give 3 out of 6 to current textbook \\
       \textbf{Utterance 2} & The current textbook gets a 3 out of 6\\
       \textbf{Utterance 3} & I think that the current textbook should be rated 3 out of 6\\
       \hline
       \textbf{DA} & BookRestaurant ( country = Honduras; facility = indoor; restaurant\_type = restaurant ) \\
       \hline
       \textbf{Utterance 1} & Book me a reservation for an indoor restaurant in Honduras \\
       \textbf{Utterance 2} & Book an indoor restaurant in Honduras \\
       \textbf{Utterance 3} & I need to book an indoor restaurant in Honduras \\
       \hline
       \rowcolor[gray]{0.95}\multicolumn{2}{c}{\textit{SC-GPT-NLU}}\\
       \hline
       \textbf{Utterance} & 2 of us want to eat at a restaurant that serves meatballs in VT \\
       \hline
       \textbf{DA} & BookRestaurant ( party\_size\_number = 2; restaurant\_type = restaurant; served\_dish = meatballs; state = VT ) \\
       \hline
       \textbf{Utterance} & Add the track to the Metal Talks Metallica playlist.\\
       \hline
       \textbf{DA} & AddToPlaylist ( music\_item = track; playlist = metal talks Metallica) \\
       \thickhline
    \end{tabular}
\caption{\label{tab:example} Example utterances generated by SC-GPT given dialogue acts (DA) and dialogue acts generated by SC-GPT-NLU given unlabelled utterances in Snips.} 
\end{table*} 

\subsection{Model Details}
\label{sec:model_exp}
\textbf{SLU Model.} For fair comparison, we use the same SLU model that is trained on the training data and the data augmented by our model and baseline systems. 

We adopt the same setting for the SLU model as in \cite{hou-da}. It has two layers of bi-directional LSTM with a hidden dimension of 200 and a dropout probability of 0.5. We choose the Adam optimizer \cite{adam} with a learning rate of 0.001. Gradients with a 2-norm greater than 5 are clipped. The best model is selected based on performances on the validation set. The number of training epochs is 50 and the batch size is 20. 


\noindent\textbf{Data augmentation.} For the Paired-Data-Only case, we modify the dialogue acts in the training split to construct around 300 additional combinations of DAs via dropping/inserting/replacing slots and values. For each dialogue act, we sample three utterances produced by SC-GPT. After filtering out utterances which do not contain all the slot-values, we collect around 500 synthetic utterances and add them into the original training split. 

We simulate the Rich-in-Ontology scenario by making the dialogue acts in the whole training set available, from which 500 dialogue acts are sampled and added to the training split. 

For the Rich-in-Utterance scenario, we sample 1,000 utterances in the training corpus and use SC-GPT-NLU to produce the most probable dialogue act. After filtering, around 500 utterance-DA pairs are added to the original training split.

\noindent\textbf{Implementation details.} Both SC-GPT and SC-GPT-NLU are finetuned for 5 epoches with a learning rate as 5e-5. Nucleus sampling \cite{nucleus} is used for decoding, where the sampling top-p is 0.9, and the temperature is 1. Details on SC-GPT including the number of parameters and pretraining procedure can be found at \cite{scgpt}. The finetuning takes about half an hour on a V100 GPU machine 64GB memory. 


\noindent\textbf{Baselines.} The baseline data augmentation systems include the seq2seq \cite{hou-da} and variational autoencoder (VAE) data augmentation model \cite{vae-da2}. We also report the results for the case without data augmentation, denoted by No-DA.

\subsection{Results}
Table~\ref{tab:main} shows the accuracy of slot tagging and intent classification for various models. Based on the results, we make the following observations.

Firstly, our data augmentation method can considerably boost the model accuracy (comparing No-DA and Ours), especially when the training data size is small. For instance, in ATIS, when only paired data is available, the slot F1 increases by 6.51 (Small) and 1.31 (Medium) points, while the intent accuracy increases by 1.68 (Small) and 0.56 (Medium) points. 

Secondly, under Rich-in-Ontology and Rich-in-Utterance scenarios, our method further boosts the slot F1 by up to 7 points and intent accuracy by up to 2.4 points. Overall, the accuracy scores are the highest when the ontology information is available. This shows that our method can take advantage of additional information and produce better synthetic training data for downstream models. We conduct statistical paired t-tests and find that the best model's performance is all statistically significant with p-value less than 0.05.

Thirdly, under the traditional Paired-Data-Only scenario, our data augmentation method outperforms all baselines in ATIS-Small, and achieves comparable results in ATIS-Medium. This shows that our method is better suited when training data is scarce.

\subsection{Examples of Augmented Data}
In Table~\ref{tab:example}, we show examples of generated utterances and dialogue acts by SC-GPT and SC-GPT-NLU in Snips. As shown, after pretraining and domain finetuning, SC-GPT can produce coherent utterances with a high variability, while covering all required intent, slots and values. SC-GPT-NLU can generate dialogue acts in the same format as input to SC-GPT, which captures the important semantic information in the input utterances. This demonstrates that pretrained models can quickly adapt to target domains with a small amount of labeled data. This facilitates the generation of high-quality synthetic data for SLU. 

\section{Conclusion}
\label{sec:conclusion}
In this paper, we approach the problem of data scarcity in SLU with pretrained language models. After finetuning on domain-specific dialogue data, our model can produce high-quality synthetic data which boosts the performance of the downstream SLU model. Moreover, we provide solutions to two semi-supervised scenarios in SLU overlooked by previous literature: \textit{Rich-in-Ontology} and \textit{Rich-in-Utterance}. In experiments on the benchmark datasets ATIS and Snips, we demonstrate that our solution can effectively leverage auxiliary unlabeled data to produce high-quality synthetic training data for building SLU models with a higher accuracy.

As future work, we aim to extend the idea of data augmentation based on pretrain language models to other speech language processing tasks, such as information retrieval and summarization.

\bibliography{main}
\bibliographystyle{IEEEtran}
\end{document}